\documentclass[letterpaper, 10 pt, conference]{ieeeconf}  

\IEEEoverridecommandlockouts                              

\usepackage{graphics} 
\usepackage{epsfig} 
\usepackage{amsmath} 
\usepackage{amssymb}  

\usepackage{cite}
\usepackage{url}
\usepackage[margin=0.75in]{geometry}
\usepackage{overpic}
\usepackage{xcolor}
\usepackage{tikz}

\title{\LARGE \bf
MaskVal: Simple but Effective Uncertainty Quantification for 6D Pose Estimation
}

\author{Philipp Quentin$^{1,2}$ and Daniel Goehring$^{2}$
\thanks{\copyright 2024 IEEE.  Personal use of this material is permitted.  Permission from IEEE must be obtained for all other uses, in any current or future media, including reprinting/republishing this material for advertising or promotional purposes, creating new collective works, for resale or redistribution to servers or lists, or reuse of any copyrighted component of this work in other works.}%
\thanks{$^{1}$BMW Group, Munich, Germany}%
\thanks{$^{2}$Freie Universitaet Berlin, Dahlem Center for Machine Learning and Robotics, Berlin, Germany }%
}

\begin{document}

\maketitle
\thispagestyle{empty}
\pagestyle{empty}

\begin{abstract}

For the use of 6D pose estimation in robotic applications, reliable poses are of utmost importance to ensure a safe, reliable and predictable operational performance. Despite these requirements, state-of-the-art 6D pose estimators often do not provide any uncertainty quantification for their pose estimates at all, or if they do, it has been shown that the uncertainty provided is only weakly correlated with the actual true error. To address this issue, we investigate a simple but effective uncertainty quantification, that we call MaskVal, which compares the pose estimates with their corresponding instance segmentations by rendering and does not require any modification of the pose estimator itself. Despite its simplicity, MaskVal significantly outperforms a state-of-the-art ensemble method on both a dataset and a robotic setup. We show that by using MaskVal, the performance of a state-of-the-art 6D pose estimator is significantly improved towards a safe and reliable operation. In addition, we propose a new and specific approach to compare and evaluate uncertainty quantification methods for 6D pose estimation in the context of robotic manipulation.

\end{abstract}


\section{Introduction}
	
6D pose estimation is an essential part of various fields such as augmented reality, mobile robots and especially manipulation robots. When such robots are used in safety-critical areas or areas that require reliable and predictable performance, as required in the manufacturing industry, a reliable uncertainty quantification (UQ) that assesses the correctness of an estimated pose is crucial. In industrial robot applications, such quantification is necessary to ensure successful grasping and to avoid collisions with the target object itself or its peripherals, thus avoiding damage and downtime. On top of that, it reduces the number of unsuccessful grasps, thus reducing overall grasp times and enables active vision strategies. Without such quantification, the deployment of vision-based manipulation robots in weakly structured environments is difficult, if not impossible.

In this context, the desired property of an uncertainty quantification is that it has a high correlation with the true underlying pose error. This allows the specification of thresholds that can guarantee a successful grasp. Furthermore, it is also desirable that while ensuring a successful grasp, not too many valid poses that would also lead to a successful grasp are discarded and thus increase the process time unnecessarily. Along with this, such an uncertainty quantification should require little or no additional computational power and process time in order to remain within industrial requirements. 

Against this background, current state-of-the-art 6D pose estimators do not provide an uncertainty quantification at all \cite{Wang_2021_CVPR, He_2021_CVPR}, or if they do, the provided uncertainties are prone to be overconfident \cite{Shi2020, Quentin2023}.

In this paper, we investigate a simple but effective uncertainty quantification for 6D pose estimation that requires no modification of the estimator itself, that we call MaskVal, which compares and validates an estimated pose by rendering with its corresponding mask detection. In the case of a common two-stage pose estimator that already relies on an instance segmentation in the first stage, the masks are already given and the additional computational effort is comparatively small. Despite MaskVal's simplicity, we demonstrate through an evaluation on a dataset and a robotic setup that MaskVal achieves state-of-the-art performance and even outperforms the recently proposed ensemble method \cite{Shi2020}. As a further contribution of the paper, we propose specific performance metrics that allow a detailed analysis and evaluation of uncertainty quantification for 6D pose estimation and by that naturally enhance the common area under the curve analysis established by \cite{2018_Xiang_RSS}.

\section{Related Work}
\label{sec:related_work}

With the deployment of neural networks in real-world applications, the need for reliable uncertainty quantification (UQ) of their predictions has risen, and different approaches have emerged. These approaches can  be categorized into those that are more suited to capture the uncertainty that stems from the noise of the data, the aleatoric uncertainty, or the uncertainty that stems from the data dependent model parametrization itself, the epistemic uncertainty \cite{2017_Kendall_NIPS, 2022_ValdenegroToro_CVPR}. 

In the field of 6D pose estimation leading and state-of-the-art models often focus mainly on the core 6D pose task itself and do not provide an uncertainty at all as in \cite{Wang_2021_CVPR, He_2021_CVPR, 2020_Labbe_ECCV, He2020_CVPR}. In contrast to that \cite{2019_Wang_CVPR, DOPE} directly provide an uncertainty for their estimation, which can be categorized as an aleatoric uncertainty, by integrating an uncertainty value directly in their training loss function that rewards the model to provide high values when the pose error is expected to be high for an given input. However, it has been shown that the uncertainties provided by this approach only weakly correlate with the actual true error \cite{Shi2020, Quentin2023}.

A comparable approach, that can also be seen as modeling aleatoric uncertainty, is also not to learn to predict single target value, but the first and second moments of a probability distribution, where the first moment is then taken as the target prediction and the second moment is the uncertainty measure. In \cite{2017_Lakshminarayanan_NIPS} they learn to predict a Gaussian distribution and in the context of 6D pose estimation in \cite{Deng2020}, they learn to predict a Bingham distribution that is especially suited to model rotational uncertainties. A disadvantage on focusing on aleatoric uncertainty alone is, that the results in \cite{2022_ValdenegroToro_CVPR}  have recently shown that aleatoric uncertainty is unreliable to find out-of-distribution data. This is especially problematic in the field of 6D pose estimation, since estimators nowadays are often trained on synthetic data only, which is prone to have a domain gap towards the encountered real data.

In contrast to these approaches where model parametrization is deterministic, Bayesian deep learning applies a probability distribution over model parameters to capture epistemic uncertainty related to data. As the full computation of the posterior of such neural networks is intractable, due to their large parameter space, Bayesian approximations like Monte-Carlo-Sampling, Dropout-Monte-Carlo-Sampling and Deep Ensembles have emerged \cite{2020_Wilson_NIPS}. Furthermore, \cite{2022_ValdenegroToro_CVPR} shows that Deep Ensembles are capable of modeling both aleatoric and epistemic uncertainty and achieve the best uncertainty behavior in their experiments. Against this background \cite{2017_Lakshminarayanan_NIPS} trains and deploys an ensemble of differently initialized neural networks with the same architecture as an approximation. A practical disadvantage of this approach is that no pre-trained models can be used as they need to be retrained. As an alternative to this, \cite{Loquercio2020} applies dropout at inference to already trained models to obtain a distribution over parameters and capture epistemic uncertainty.

In the specific context of 6D pose estimation, \cite{Shi2020} uses an ensemble of neural networks to model uncertainty, where the models have different architectures and have also been trained on different datasets. Uncertainty is then quantified by computing the ADD error \cite{2012_Hinterstoisser_ACCV} between the different ensemble poses or by a learned metric. 

A general drawback of these approaches is that they require the inference of multiple networks, which can be done in parallel, but still require higher computational
resources, and may conflict with industry constraints.

Against this background, we investigate a computationally
inexpensive approach that does not require any modification of the pose estimators, by comparing corresponding instance segmentations and rendered pose estimations in 2D image space. In doing so it can as well be interpreted as an ensemble approach, but on segmentation instead on pose level. Since the majority of state-of-the-art pose estimators already consist of an instance segmentation in their first stage, the existing instance segmentation can be used and no further computational effort is needed for an ensemble. 

Our method can be categorized as a special case of the general render-and-compare approach, which is well established in the evaluation of object pose estimation.\cite{Deng2021}, for example, compares the cosine distance of image features of the rendered object with observed object image features via an autoencoder architecture to evaluate pose candidates. \cite{Gard2022}, on the other hand, uses a render-and-compare approach to access the alignment of rendered object contours and image edges for pose validation. Closest to our investigated approach, the works of \cite{2020_Wang_ECCV, Wang2021} also compare the instance segmentations with their corresponding pose estimates by rendering in the 2D image space. However, their focus is on self-supervised retraining and, to the best of our knowledge, no other work explicitly addresses this approach for direct pose uncertainty quantification as we investigate here.

\section{Problem Formulation and Evaluation Methodology}
\label{sec:problem}

In the following, we describe the problem setting and propose a methodology to evaluate the properties of an uncertainty quantification method for 6D pose estimation for industrial robotic applications. 

Consider the goal of a 6D pose estimation system to provide poses $\hat{p}_{(i)}$ for target objects that enable a robot to successfully grasp and place them. Therefore, suppose  a scene of known objects with an associated set of ground truth poses $ O = \{\bar{p}_{(1)}, \dots, \bar{p}_{(O)} \}$ that are to be grasped by the robot. The 6D pose estimator provides a set of pose estimations  $ P = \{(\hat{p}_{(1)}, u_{(1)}), \dots, (\hat{p}_{(N)},  u_{(N)}) \}$   with corresponding uncertainties $u_{(i)}$. The uncertainties shall enable the pose estimator to reject pose estimations which are not accurate enough for a successful grasp (invalid) and provide a set of filtered poses $ P_{u} = \{\hat{p}_{(1)}, \dots, \hat{p}_{(M)} \}$ based on an uncertainty threshold $u_{T}$. In this context, an optimal uncertainty quantification enables the objective that the filtered set contains only valid poses, but at the same time no valid poses of the pose estimator are discarded.

To evaluate towards this objective, we first compute a pose error by a pose error function $e = E(\bar{p}, \hat{p})$, that enables the categorization, if a pose will lead to a successful grasp or not for a specific error threshold $e_{t}$. Furthermore, this enables a general assessment of the uncertainty quantification by computing the correlation between the uncertainties and the true error $e$ as in \cite{Shi2020}. Further, we define a true positive (TP) for the case when
\begin{equation}
 E(\hat{p}_{(i)}, \bar{p}_{(j)}, ) \leq e_{t}
\label{eq:tp}
\end{equation}
a false positive (FP) when
\begin{equation}
 E( \hat{p}_{(i)}, \bar{p}_{(j)}) > e_{t}
\label{eq:fp}
\end{equation}
and a false negative (FN) for each case where there is no or no true positive pose estimation for a ground truth pose. For the sake of simplicity, we assume at this point that the association between $i$-th estimated pose and the $j$-th ground truth pose is known. Given this categorization, we can now compute the common performance metrics average precision (AP) as
\begin{equation}
AP = \frac{\#_{TP}(P_{u})}{\#_{TP}(P_{u}) + \#_{FP}(P_{u})}
\label{eq:apg}
\end{equation}
and the average recall (AR) as
\begin{equation}
AR = \frac{\#_{TP}(P_{u})}{\#_{TP}(P_{u}) + \#_{FN}(O)}
\label{eq:arg}
\end{equation}
where $\#(\cdot)$ returns the number of TPs, FPs and FNs respectively with regard to a specific pose set. To additionally measure how many actual true positive poses in $P$ were not rejected, we further compute the ratio of true positives, which we call the average recall uncertainty (ARU), in $P_{u}$ and $P $ as
\begin{equation}
ARU = \frac{\#_{TP}(P_{u})}{\#_{TP}(P)}.
\label{eq:arug}
\end{equation} 
For our initially supposed objective of an optimal uncertainty quantification for a given pose estimator, an optimal uncertainty quantification would enable an AP of 1 and simultaneously an ARU of 1 for a specific error threshold. By that the provided set of estimated poses of the pose estimator is best possibly used.

\section{MaskVal}
\label{sec:MaskVal}

To create an uncertainty respectivley certainty measure for a pose estimation of an object of interest, we compare the pose with its corresponding instance segmentation, by rendering the object's model transformed by the pose into the 2D image plane and compute the corresponding intersection over union (IOU). Specifically, given an RGB or RGB-D image $I^{w \times h}$, with width $w$ and height $h$ $\in \mathbb{N} $, a pose estimator $g$ provides a set of 6D poses $ P = \{\hat{ p}_{(1)}, \dots, \hat{p}_{(N)} \}$ and an instance segmentation network $s$ provides a set of  instance segmentations $ S = \{ is_{(1)}, \dots, is_{(K)} \}$, where $is \in \{0, 1\}^{w \times h}$ is a binary mask. Furthermore, given a renderer $ r(\hat{p}_{(i)}, M_{(i)}, K, w, h) =  (D_{(i)}, v_{(i)}) $ where $ D \in \mathbb{R}_{\geq 0}^{w \times h} $ and $v \in [0,1]$, which renders the depth of the target object model  $M_{(i)}$ transformed by the pose $\hat{p}_{(i)}$  to the 2D image space with camera matrix $K$ and provides a visibility ratio $v_{(i)}$ that reflects the ratio of how much of the object is visible in the image due to field of view (FoV) limitations. Then we can compute the pairwise mask IOUs for the $N$ poses and $K$ instance segmentations as
\begin{equation}
iou_{ik} = \frac{ \sum_{n=1}^{w} \sum_{m=1}^{h}(  \delta_{1}(D_{(i)}) \odot is_{(k)})_{nm}}{  \sum_{n=1}^{w} \sum_{m=1}^{h}(\delta_{1}(D_{(i)})  \oplus is_{(k)} )_{nm} }  
\label{eq:iou}
\end{equation} 
where we define $\odot$ as the elemt-wise AND operator,  $\oplus$  as the elemt-wise OR operator and 
\begin{equation} 
\delta_{1}(D_{(i)}) = \mathbf{1}_{\{D_{nm} > 0\}}
\label{eq:indicator}
\end{equation}
is an indicator function that maps an input matrix to an binary output matrix, where an entry is 1 if  $D_{nm}$ is greater than 0 and 0 otherwise. 

Using finally a matching algorithm $J$ that takes as input the queried pose index $i$ and the  $ iou \in [0,1]^{N \times K} $ matrix to establish the most likely pose instance segmentation association, we choose the pose certainty to be
\begin{equation} 
J(i, iou) = iou_{ik^{*}} = c_{(i)}
\label{eq:pu}
\end{equation}
with $c_{(i)} \in [0,1]$ and where $J$ can be realized by a greedy algorithm, for example. The corresponding uncertainty $u_{(i)}$ can then simply be retrieved by the relation 
\begin{equation} 
u_{(i)} = 1 - c_{(i)}.
\label{eq:relation_u_c}
\end{equation}

 For the common case of a two-stage pose estimator, where the pose estimation  builds directly up on the instance segmentation, the association is known, the matching algorithm $J$ can be omitted and the certainty simplifies to
\begin{equation} 
 iou_{(ik; i=k)} = c_{(i)}.
\label{eq:pue}
\end{equation}

To account for the case where the object is barely visible in the image due to FoV limitations, we assume that the mask IOU is less reliable as an uncertainty quantification and therefore we decrease the certainty by the visibility ratio and get the uncertainty as

\begin{equation} 
u_{(i)}= \begin{cases}
1 - c_{(i)} v_{(i)} & v_{(i)} < \alpha \\
1 - c_{(i)} & \, v_{(i)} \geq \alpha
\end{cases}
\label{eq:vis}
\end{equation}
where $\alpha \in [0,1]$ sets the sensibility towards the visibility. 

\begin{figure*}[t!]
\centering
\setlength{\fboxsep}{0pt}
\setlength{\fboxrule}{0pt}
\framebox{\parbox{7in}{
\centering
\begin{tabular}{ccc}
\includegraphics[width=0.3\linewidth]{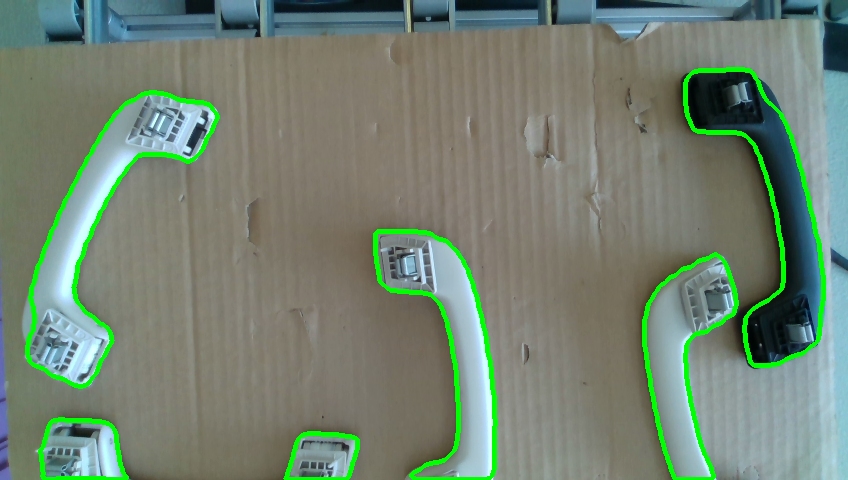} &
\includegraphics[width=0.3\linewidth]{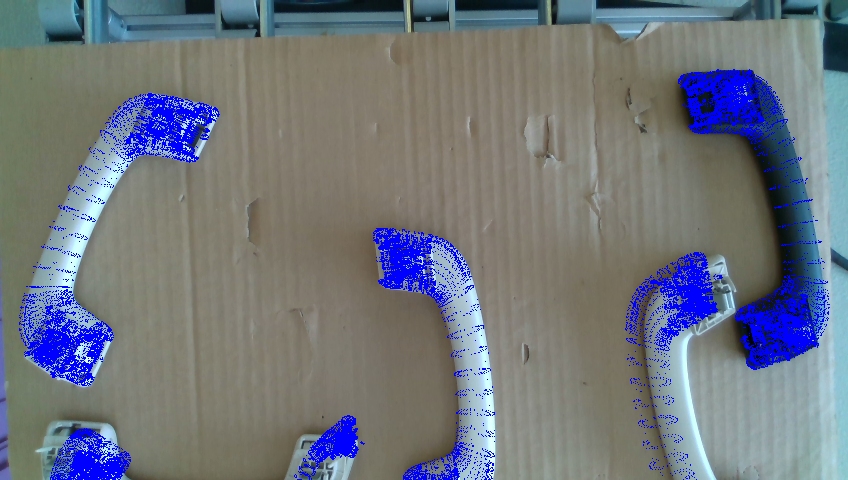}  & \includegraphics[width=0.3\linewidth]{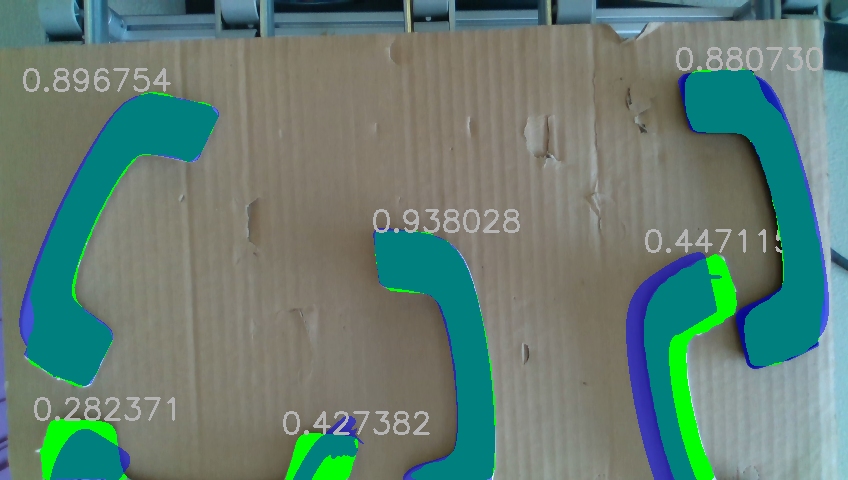} \\
\end{tabular}
}}
\caption{In the figure, from left to right: (1) Object detection with instance segmentation of Mask R-CNN, (2) the projected handle object model in transformed poses of GDR-Net, and (3) comparison of instance segmentations with projected pose masks accompanied by the corresponding certainty values of MaskVal incorporating the visibility ratios.}
\label{Fig:MaskVal}
\end{figure*}

To get a visual impression of MaskVal, we depict an exemplary uncertainty quantification for a data sample of the robotic experiments from section \ref{sec:Experiments}  in Fig. \ref{Fig:MaskVal}.

\section{Experiments}
\label{sec:Experiments}

In the experiment section we evaluate and compare our method MaskVal against the recent state-of-the-art uncertainty quantification method for 6D poses from \cite{Shi2020} on our objectives from section \ref{sec:problem}.  We conduct the evaluation and comparison on a test dataset with known ground truth as well as on a robotic setup for two representative parts of the automotive internal logistics, namely antennas and handles, whose synthetic versions are depicted in Fig. \ref{Fig:synthetic_data}. In this context, the antennas represent a rather difficult part due to  their reflective surface and the absence of well distinctive features, whereas the handles represent a rather common simple part.

\subsection{Datasets}
\label{sec:Datasets}

We use only synthetic data for the training of all algorithms used, as it is one of the most interesting scenario for industry \cite{Quentin2023}.

Furthermore, we also only use synthetic data for our dataset evaluation part  for the uncertainty quantification methods. The reason for this is, that it cannot be guaranteed, that real annotated pose data do not contain errors in a range of values, which is essential and part of our evaluation. This makes a fine-grained evaluation of the uncertainty quantification methods difficult, as errors may stem not from the method itself, but from underlying annotation errors. Therefore, we have only used synthetic data to exclude any such unfavorable effects. The experiments on the robotic setup shall compensate the lack of real data at this point.

\begin{figure*}[t!]
\centering
\setlength{\fboxsep}{0pt}
\setlength{\fboxrule}{0pt}
\framebox{\parbox{7in}{
\centering
\begin{tabular}{cccc}
\includegraphics[width=0.22\linewidth]{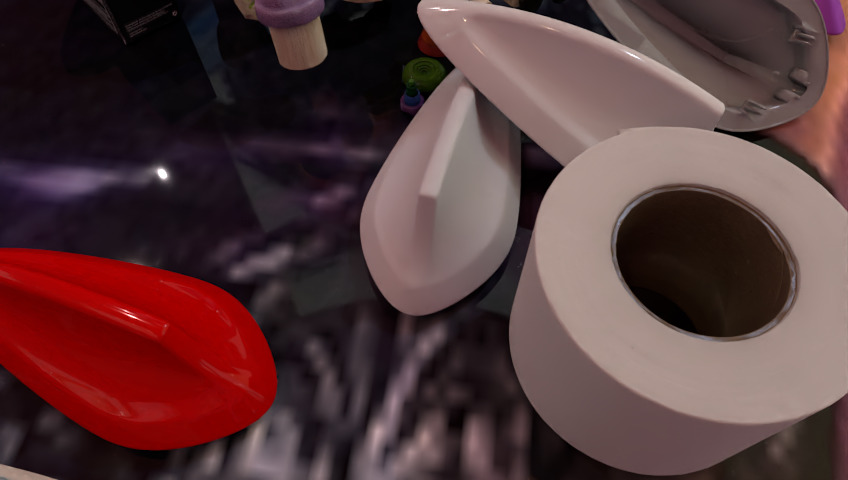} &
\includegraphics[width=0.22\linewidth]{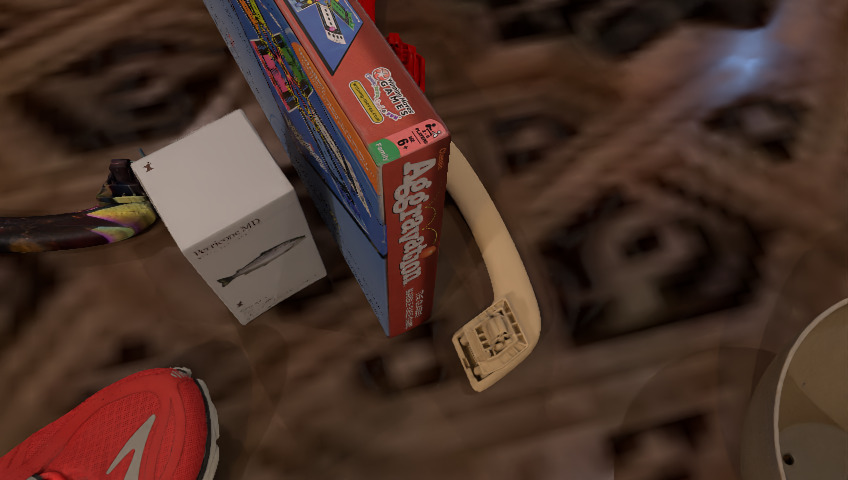}  & \includegraphics[width=0.22\linewidth]{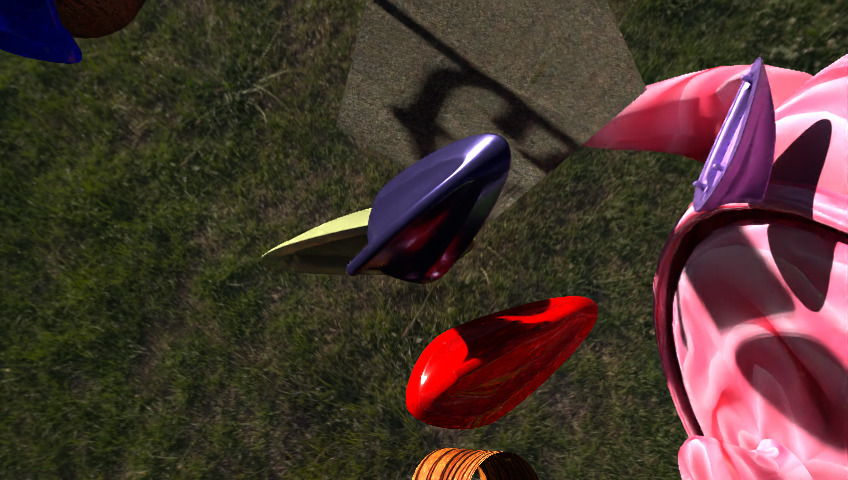} & \includegraphics[width=0.22\linewidth]{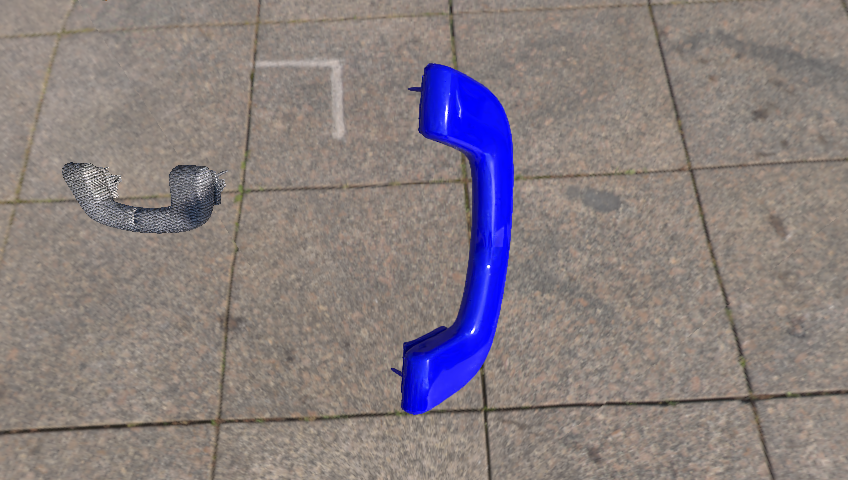} \\
\end{tabular}
}}
\caption{In the figure, from left to right:  photorealistic lightweight scenes (S1) from the test dataset of the antenna (1) and handle (2) and render \& paste scenes (S2) of the antenna (3) and handle (4).}
\label{Fig:synthetic_data}
\end{figure*}

For the generation of the synthetic data, we have used the NVISII-based data generation pipeline of \cite{Quentin2023}. Using this pipeline, we have created two training and validations data sets, S1 and S2, as well as an overall test dataset for each part, as specified in Tab. \ref{tab1}.
\begin{table}[t]
\caption{Datasets}
\begin{center}
\begin{tabular}{|c|c|c|c|}
\hline
\textbf{Name} & \textbf{Train} & \textbf{Validation}& \textbf{Test} \\
\hline
 \begin{tabular}{@{}c@{}}S1 \\ S2\end{tabular} & \begin{tabular}{@{}c@{}} 50k synth. img \\ 50k synth. img \end{tabular} & 
    \begin{tabular}{@{}c@{}} 10k synth. img. \\  10k synth. img.\end{tabular}
  & 10k synth. img. \\
  \hline
\end{tabular}
\label{tab1}
\end{center}
\end{table}
The training and validation dataset S1 and the overall test data set are created by the photorealistic lightweight synthetic scenes of \cite{Quentin2023}. The training and validation dataset S2 differs from S1 in that the objects are not spawned in a physical room with a physical background, but in front of a 2D image, which is the common render \& paste approach \cite{Hodan2020}. To evaluate the uncertainty quantification methods on various conditions, the test dataset contains heterogenous objects with clutter and occlusions as well as random textures on the target objects. Exemplary images of the datasets are depicted in Fig. \ref{Fig:synthetic_data}.

\subsection{Implementation Details}

\subsubsection{6D Pose Estimation}

For the underlying 6D pose estimation we use the state-of-the art two-stage RGB-based pose estimator GDR-Net \cite{Wang_2021_CVPR}, which is the winner of the BOP Challenge 2022 \cite{2023_Sundermeyer_CVPR}. For the first stage of GDR-Net, the instance segmentation, we use the basic Mask R-CNN algorithm \cite{2017_He_ICCV}. Both GDR-Net and Mask R-CNN were trained on  dataset S1. Note that GDR-Net by itself does not provide an uncertainty quantification for its pose estimations.

\subsubsection{Uncertainty Quantification via Pose Ensemble}

We benchmark our method against the recent state-of-the-art uncertainty quantification method from \cite{Shi2020}, which we call Ensemble-ADD, that uses a heterogeneous ensemble of 6D pose estimators to provide an uncertainty via a disagreement metric. Following their approach, we selected their well-performing ADD-based disagreement metric for our comparison and obtained a heterogeneous ensemble by training a second GDR-Net on a different dataset, namely dataset S2. Consistent with our approach, the GDR-Net that was trained on dataset S1 provides the final pose estimations. Since the ADD-based disagreement metric is unbounded, we normalize the obtained ADD disagreement for better comparability to obtain the uncertainty $u$ in an interval of $[0,1]$. To do this, we choose the normalization minimum as the minimum ADD disagreement obtained on the test set and set the normalization maximum to 5 cm to exclude high outliers.

\subsubsection{MaskVal}
Since GDR-Net is a two-stage pose estimator that already relies on an instance segmentation in the first stage, we can directly compare the poses with their corresponding instance segmentation provided by Mask R-CNN as described in section \ref{sec:MaskVal}. For the renderer we use the off-screen C++ renderer of the BOP-Toolkit, for fast CPU-based rendering. Furthermore, we set the parameter $\alpha = 0.8$, which means that the visibility ratio will be taken into account, when it drops below this value. 

\subsection{Metric Details for Dataset Evaluation}

For realizing the general metric description of section \ref{sec:problem} for the dataset evaluation, we follow \cite{Quentin2023} and define $E(\hat{p}_{(i)}, \bar{p}_{(i)})$ as the maximum distance error (MDD) as
\begin{equation}
E_{MDD}(\hat{p}_{(i)}, \bar{p}_{(i)}, \mathcal{M}_{(i)} )=\underset{x \in \mathcal{M}_{(i)}}{\operatorname{max}}\|\hat{p}_{(i)} x- \bar{p}_{(i)} x \|_{2}
\end{equation} 
where a pose $p$ is defined as a homogeneous transformation matrix consisting of a rotation matrix $R \in SO(3)$ and a translation vector $t \in \mathbb{R}^{3}$ and $\mathcal{M}_{i}  \in \mathbb{R}^{N \times 3} $ is the target object's model point cloud. 

For the realization of the performance metrics we also follow \cite{Quentin2023} and specify
\begin{equation}
AP = \frac{1}{N} \sum_{n=1}^{N} \frac{TP_{u,n}}{TP_{u,n}+FP_{u,n}} \label{eq:ap}
\end{equation}
\begin{equation}
AR =  \frac{1}{N} \sum_{n=1}^{N} \frac{TP_{u,n}}{TP_{u,n}+FN_{n}} \label{eq:ar}
\end{equation}
\begin{equation}
ARU =  \frac{1}{N} \sum_{n=1}^{N} \frac{TP_{u,n}}{TP_{n}} \label{eq:aru}
\end{equation}
where $TP_{u,n}$, $FP_{u,n}$ and $FN_{n}$  is the number of TPs, FPs and FNs in the filtered set of estimated poses for an image $n$ of dataset $N$ and $TP_{n}$, is the number of TPs regarding the unfiltered set of poses, respectively. For the specific definition of a TP, FP and FN in image datasets, we follow \cite{Quentin2023} and also set the visibility threshold to $\theta_{v} $ = 0.85. This means that objects are only counted as a false negative, if there is no provided pose estimation and simultaneously the object's visibility is equal or greater than 85 \%. Thereby the pose estimator is only penalised by FNs for objects for which it should definitely be able to provide a valid pose.

\subsection{Results on Dataset}

The results of Mask R-CNN on the test are depicted in Tab.\ref{tab2}. Since  Mask R-CNN was trained and tested on the same synthetic data domain, the results are good as expected. 
\begin{table}[t]
\caption{Mask R-CNN Results }
\begin{center}
\begin{tabular}{|c|c|c|c|}
\hline
\textbf{Object} & \textbf{$\mathbf{AP_{50:95(\%)}}$}& $\mathbf{AP_{50(\%)}}$ & \textbf{$\mathbf{AR^{100}}$}  \\
\hline
Antenna    & 94.7 & 98.9 & 96.8 \\
Handle      & 87.3 & 99.0 &  90.4 \\
\hline
\end{tabular}
\label{tab2}
\end{center}
\end{table}

\begin{figure}[tb]
\centering
\setlength{\fboxsep}{0pt}
\setlength{\fboxrule}{0pt}
\framebox{\parbox{3in}{\centerline{\includegraphics[width=0.49\textwidth]{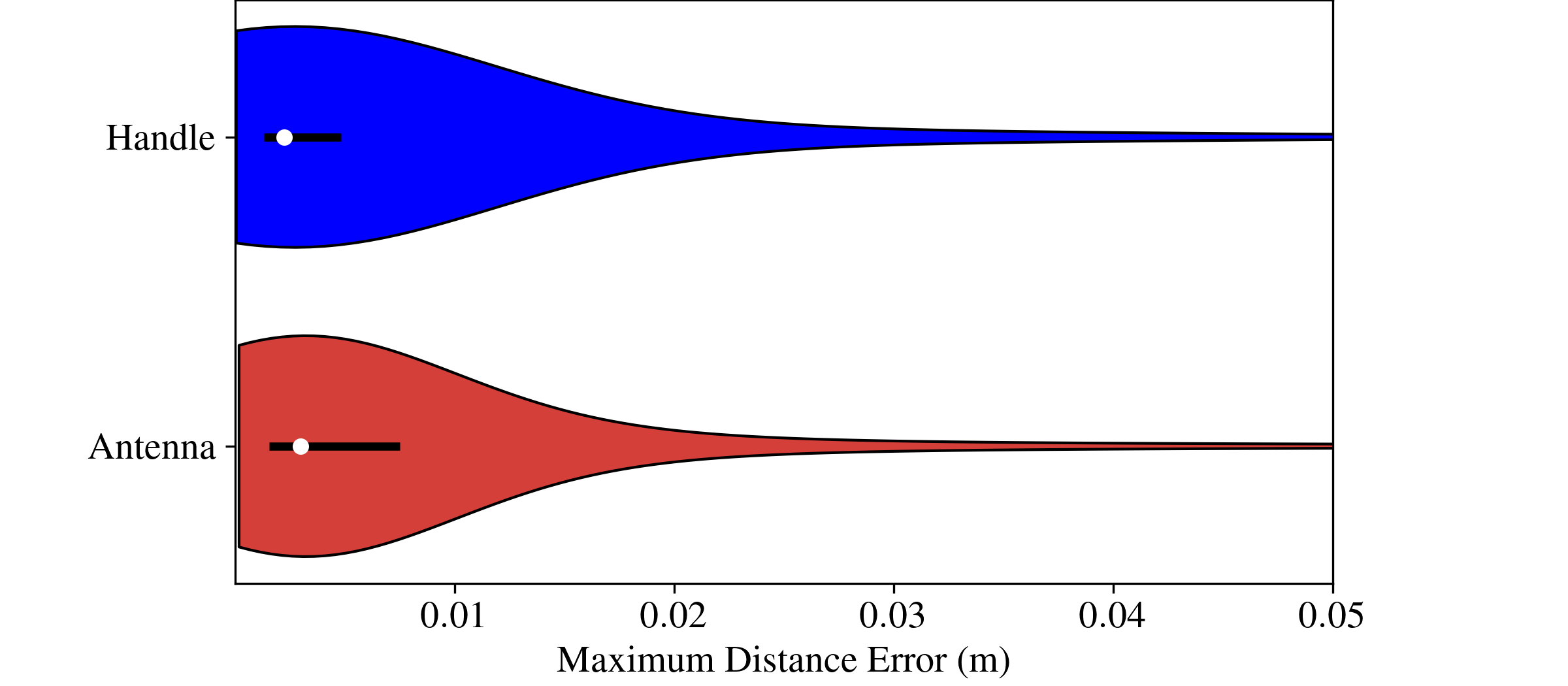}}
}}    
\caption{Violin plots of the MDE on the test set for the for the antenna and handle. The white dots within the violin plots represent the location of the median and the black bars indicate the range of the lower and upper quartiles, that enclose 75 \% of the data.}
\label{violinplot}
\end{figure}

To get an impression of the overall performance of GDR-Net trained on S1, we show the pose error distribution on the test set for the antenna and handle in a violin plot in Fig. \ref{violinplot}. Despite that the majority of pose errors are quite small and will likely represent a successful grasp, there is nevertheless a significant amount of poses that will likley lead to an unsuccessful grasp, including collisions with the target object or the environment. As a solution for an industry deployment such an error distribution is insufficient. Depending on the specific task and gripper combination, that only allows for a certain maximum error, the goal is to reliably discard all poses with an error greater than this, while keeping the poses with a smaller error. 

\begin{figure*}[tb]
\begin{center}
\begin{tabular}{cc}
\includegraphics[width=0.45\linewidth]{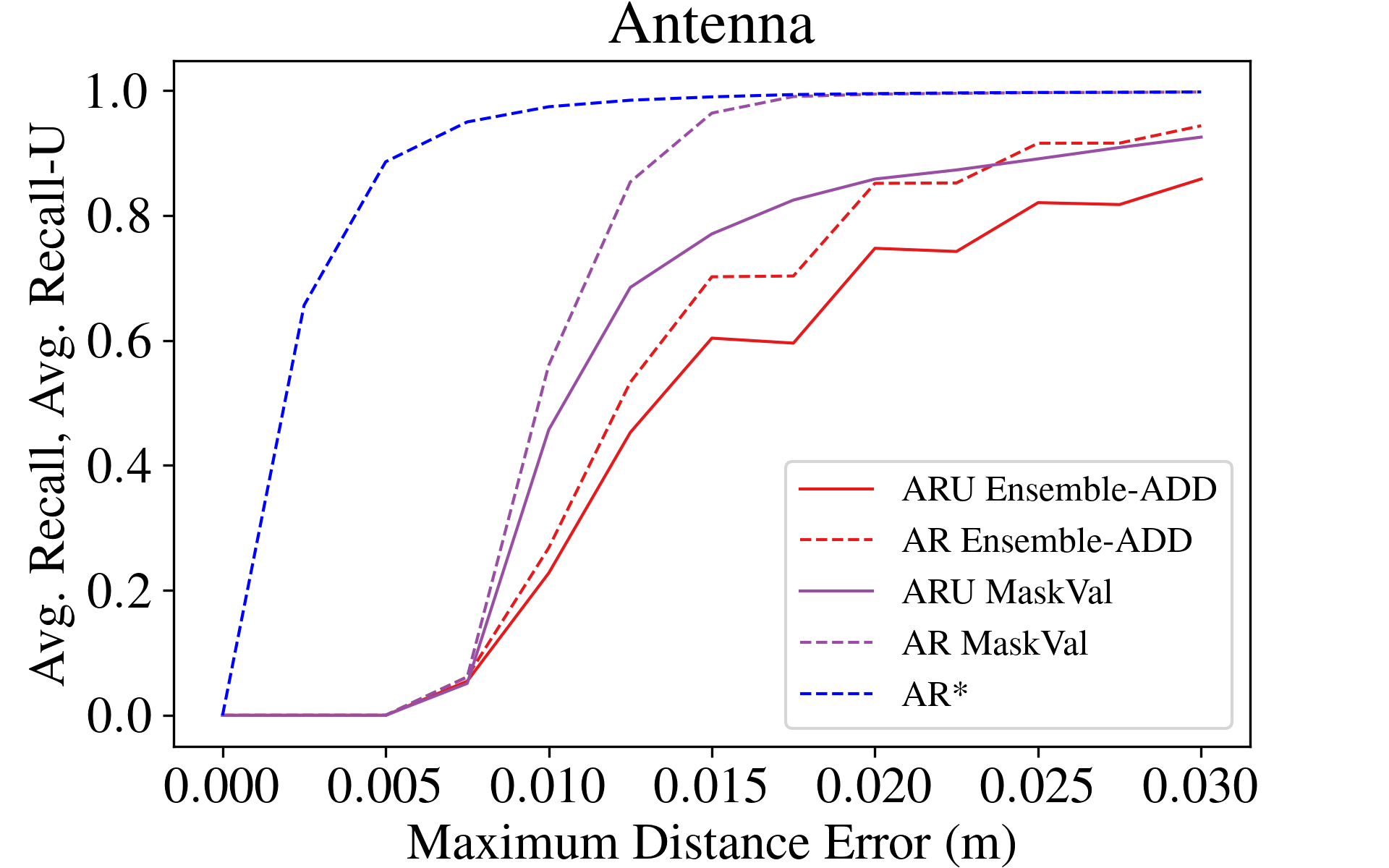} &
\includegraphics[width=0.45\linewidth]{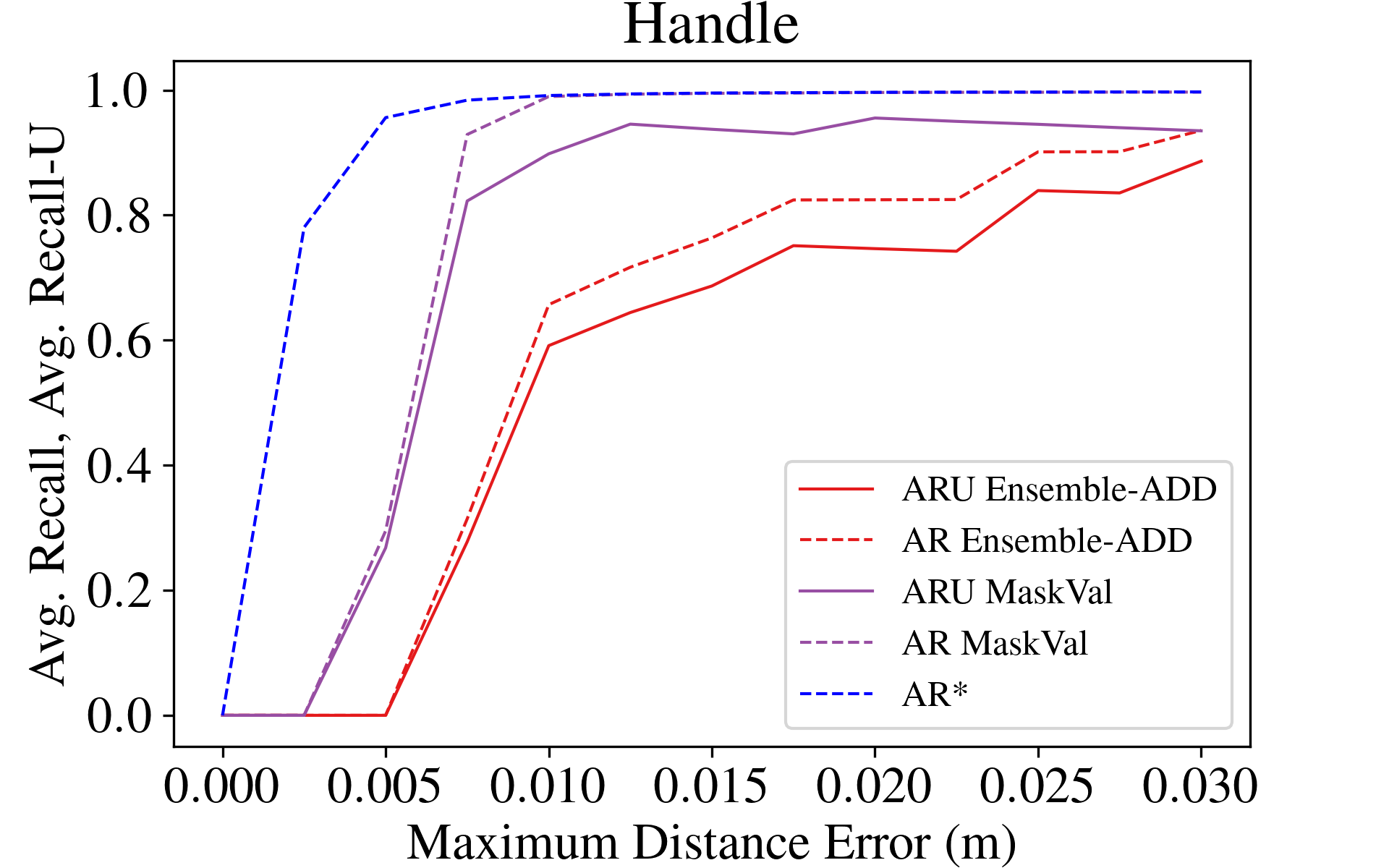} \\
\end{tabular}
\end{center}
\caption{Average recall (AR) and average recall uncertainty (ARU) curves for the antenna and handle on the test dataset over the MDD error threshold $e_{t} \in [0, 0.03]$ m, such that the average precision is equal to or greater than 0.99 along the entire curves. The AR* curve is the average recall curve of the unfiltered pose set, for which the above average precision condition does not apply.}
\label{Fig:Graphs}
\end{figure*}

In Fig. \ref{Fig:Graphs} we show how well the uncertainty methods MaskVal and Ensemble-ADD are able to achieve this goal. Specifically, based on their provided uncertainties for the pose estimates, we have set the uncertainty threshold such that the AP of equation (\ref{eq:ap}) is equal to or greater than 0.99 for each error threshold $ e_{t} \in [0, 0.03]$ m and plotted the corresponding AR and ARU curves. This means that along the entire AR and ARU curve, the AP is 0.99, which means that only 1 \% of the filtered poses have a larger error. We chose an AP of 0.99 because it is a realistic assumption for industrial applications, but higher values may be required. In addition, we also plotted the average recall curve AR* of the unfiltered pose set, as commonly done since \cite{2018_Xiang_RSS}. On this curve, the AP is much lower than 0.99, but it marks the optimum that the plotted AR curve could reach with an optimal uncertainty quantification method. In other words, if the AR curve were to reach the AR* curve, it would mean that no valid poses would be rejected while simultaneously having an AP of 0.99. Therefore, the provided pose set would have been best exploited by the uncertainty quantification method with respect to a given AP target.
 
To summarize the performance of the uncertainty methods on these curves by a scalar value, we additionally compute the area under the curve (AUC) for AR and tabulate the results in Tab. \ref{tab_spearman}. Furthermore, we provide the Spearman's rank correlation between the true error and the uncertainties as in \cite{Shi2020}. The Spearman's rank correlation is able to measure also non-linear correlations and a perfect correlation or inverse correlation is given by $+1$ or $-1$ respectively.

Regarding the results presented, the first thing to note is that it is possible with both MaskVal and the Ensemble-ADD  to filter the poses in such a way that an AP of 0.99 can be achieved. This means that the uncertainties calculated by the methods correlate sufficiently with the actual true error to reliably filter out insufficient poses. In this context, the plots also show that MaskVal performs significantly better in that the uncertainty values correlate better with the true error, so that less valid poses are discarded and the AR curves reach higher values earlier. These results are consistent with the Spearman's rank correlation results in Tab. \ref{tab_spearman}.
\begin{table}[tb]
\caption{Spearman's Rank Correlation \& AUC AR}
\begin{center}
\begin{tabular}{|c|c|c|c|}
\hline
\textbf{Object} & \textbf{Method}& \textbf{Spearman's Rank Corr.} &  \textbf{AUC AR} \\
 \cline{1-4}
Antenna & \begin{tabular}{@{}c@{}} MaskVal \\ Ensemble-ADD\end{tabular}  & \begin{tabular}{@{}c@{}} \textbf{0.72} \\ 0.58 \end{tabular} & \begin{tabular}{@{}c@{}} \textbf{66.0} \\ 52.3 \end{tabular} \\
\cline{1-4}
Handle  & \begin{tabular}{@{}c@{}}MaskVal \\ Ensemble-ADD\end{tabular}  & \begin{tabular}{@{}c@{}} \textbf{0.70} \\ 0.56 \end{tabular} & \begin{tabular}{@{}c@{}} \textbf{81.0} \\ 60.0 \end{tabular} \\
\hline
\end{tabular}
\label{tab_spearman}
\end{center}
\end{table}

Furthermore, it can be noted that the AR curve based on MaskVal achieves optimal values within an error range that is sufficient for corresponding object-gripper combinations for many applications. This has direct practical implications, as it means that the same pose estimator can or cannot be used for industrial applications with specific requirements for accuracy, reliability and cycle times, depending on the uncertainty quantification method used. For example, for the antenna use case, which requires a pose accuracy of approximately 0.015 m, MaskVal allows near-optimal exploitation of the pose set, promising significantly shorter cycle times than Ensemble-ADD.
Note further that optimal AR values are reached before optimal ARU values, because false negatives are not counted for objects in the image with the set visibility threshold $\theta_{v} < $ 85 \%.

\subsection{Results on Robotic Experiment} 

\begin{figure}[t!]
\vspace*{1mm} 
    \centering
    \begin{overpic}[width=2.6 in]{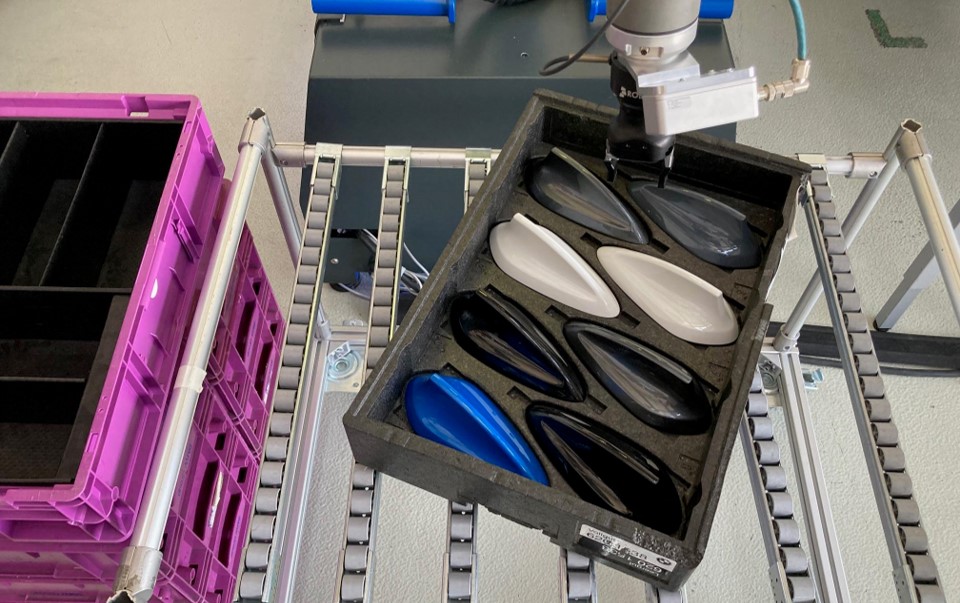}

\put (36.5,50) {\tikz \node[fill=white,inner sep=1pt] {\parbox{0.4in}{\centering\scriptsize\color{black} Camera}};}

\put (53,50) {\tikz \draw[->,line width=1pt] (0,0) -- (1, -0.1);}

\put (7,54) {\tikz \node[fill=white,inner sep=1pt] {\parbox{0.4in}{\centering\scriptsize\color{black}Sequence\\Container}};}

\put (15,50) {\tikz \draw[->,line width=1pt] (0,0) -- (0,-0.26);}

\put (85.5,34) {\tikz \node[fill=white,inner sep=1pt] {\parbox{0.3in}{\centering\scriptsize\color{black}Gripper}};}

\put (70,36) {\tikz \draw[->,line width=1pt] (0,0) -- (-1, 0.4);}

\put (88,55) {\tikz \node[fill=white,inner sep=1pt] {\parbox{0.25in}{\centering\scriptsize\color{black}UR10}};}

\put (77,56) {\tikz \draw[->,line width=1pt] (0,0) -- (-0.7, 0.3);}

\put (79,8) {\tikz \draw[->,line width=1pt] (0,0) -- (-0.4, 0.7);}

\put (76,5) {\tikz \node[fill=white,inner sep=1pt] {\parbox{0.4in}{\centering\scriptsize\color{black}Antennas}};}

    \end{overpic}
    \caption{Robotic setup representing an automotive sequencing process of internal logistics, showing a UR10 equipped with a Framos camera and a gripper. The goal is to sequence the antennas on the conveyor belt to the sequence container on the left.}
    \label{Fig:robotic_setup}
\end{figure}
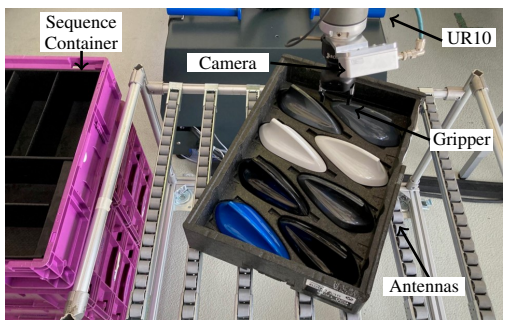

To evaluate how the results on the dataset transfer to the real world, we conducted tests on a robotic setup, depicted in Fig. \ref{Fig:robotic_setup}, which is the same as in \cite{Quentin2023}.
The setup represents a sequencing process of the internal logistics, where the goal of a robot is to pick objects from a conveyor belt and place them in a nearby sequence container. In this context, we count a TP if the robot is successful, a FP if the pose leads to a collision or an unsuccessful placing and a FN for all parts that remain on the conveyor belt after a time limit of 20 seconds per part. For more details on the setup, see \cite{Quentin2023}.

To perform the task, the robot performs a search run over the parts to be grasped. If an object is detected and a pose is estimated, the robot moves to the pose for grasping only if the uncertainty is equal or smaller than the specified pose uncertainty threshold $u_{t,g}$. If this uncertainty threshold is not reached, but the uncertainty is smaller or equal to a refinement threshold $u_{t,r}$, the robot moves to a refinement position based on the pose and captures an image of the object again. If the uncertainty is above both the grasping and the refinement uncertainty thresholds, the robot continues its original search.

In total, we performed 5 experiments per object and uncertainty method, where the robot had to sequence 8 objects of the same class per experiment. We deduced the corresponding grasping uncertainty thresholds $u_{t,g}$ from the results of the synthetic test dataset, so that for an error threshold $e_{t} = 0.015 $ m an AP of 0.99 is achieved. In this context, we obtained for the antenna a value of $u_{t,g} = 0.2$ for MaskVal and $u_{t,g} = 0.1$ for Ensemble-ADD and for the handle a value of $u_{t,g} = 0.55$ for MaskVal and $u_{t,g} = 0.08$ for Ensemble-ADD. Along with this we have set the upper uncertainty refinement threshold $u_{t,r}$ to 0.6 for all variants. In addition, we ran a subset of the experiments without any uncertainty quantification for comparison. The results of the experiments are shown in Tab. \ref{tab:robotic_experiments}.
\begin{table}[tb]
\caption{Avg. Precision \& Avg. Recall Robotic Experiment Results}
\begin{center}
\begin{tabular}{|c|c|c|c|c|}
\hline
\textbf{Object} & \textbf{Score}& \textbf{w/o UQ} &  \textbf{Ensemble-ADD} & \textbf{MaskVal}  \\
 \cline{1-5}
Antenna & \begin{tabular}{@{}c@{}} AP \\ AR\end{tabular}  & \begin{tabular}{@{}c@{}} 90.0 \\ 82.5 \end{tabular} & \begin{tabular}{@{}c@{}} 96.0 \\ 50.0 \end{tabular} & \begin{tabular}{@{}c@{}} \textbf{97.5}\\ \textbf{100} \end{tabular} \\ 
\cline{1-5}
Handle  & \begin{tabular}{@{}c@{}} AP \\ AR \end{tabular}  & \begin{tabular}{@{}c@{}} 61.4 \\ 50.0 \end{tabular} & \begin{tabular}{@{}c@{}} 100 \\ 70.0 \end{tabular} & \begin{tabular}{@{}c@{}} \textbf{100} \\ \textbf{72.5} \end{tabular} \\ 
\hline
\end{tabular}
\label{tab:robotic_experiments}
\end{center}
\end{table}

As a main result it can be stated that both uncertainty methods improved the overall performance compared to the plain GDR-Net. Furthermore, MaskVal achieved significantly better results than Ensemble-ADD and is close to optimal values for the antenna. Note that MaskVal, despite discarding pose estimates, also improved the AR performance, while achieving near-optimal and optimal results on AP against the plain GDR-Net. We observed that the reason for this is that the robot with plain GDR-Net had many time consuming paths where the gripper reached into the void, which was not the case with MaskVal.

\subsection{Interpretation of the Results}
\label{sec:discussion}

The basic idea of an ensemble method to provide an uncertainty quantification for 6D pose estimation is to relate the similarity of the ensemble outputs to the underlying pose error. The higher the similarity, the lower the uncertainty and vice versa. In this context, Ensemble-ADD tries to relate the similarity of two poses to the true underlying error, while MaskVal tries to do the same with the similarity of two instance segmentations. An essential condition for this approach to work well is that the comparison ensemble part is as good as possible, if the ensemble part that provides the actual result performs well. This generally implies that the comparison ensemble part should be as good as possible. If the comparison part were an oracle, one could theoretically infer the actual error directly. From this consideration it also follows that Ensemble-ADD can theoretically produce better results than MaskVal, since MaskVal is not able to infer the exact true pose error even with an oracle instance segmentation. However, since instance segmentation can be considered as less complex than 6D pose estimation, the results support the assumption that in practice it is easier to find a superior instance segmentation and the above mentioned disadvantage of information loss is relativized.

\section{Conclusion}
\label{sec:conclusion}

In this work we propose MaskVal as an uncertainty quantification method for 6D pose estimation for known objects, that creates an uncertainty for a pose estimate by comparing the pose with its corresponding instance segmentation in an ensemble-like manner. We show that MaskVal outperforms a previous state-of-the-art  ensemble-based method on both a dataset and a robotic setup. We further show that it enhances the performances of the state of the art RGB-based 6D pose estimator GDR-Net trained only on very lightweight synthetic data such that a deployment to industrial sequencing process is feasible. This is an important implication, as the combination of lightweight synthetic data with an RGB-based estimator is very promising in terms of economic scalability for industrial use cases.
Furthermore, our work implies that the direct uncertainty quantification of two-stage pose estimators can be significantly improved, since the mask and the pose are already inherently available in such estimators.
As a conclusion, our results generally suggest a stronger focus on uncertainty quantification in the field of 6D pose estimation, on the one hand on the estimators themselves and on the other hand in established benchmarks.

\section*{ACKNOWLEDGMENT}

The authors would like to thank Dino Knoll for the fruitful discussions and the Logistics Robotics Team at BMW and especially Jan Butz for their support. We used DeepL for light-editing such as for minor translations, spelling and grammar corrections.





\bibliographystyle{IEEEtran}
\bibliography{IEEEabrv,../../jabref/dissertation.bib}

\end{document}